\documentclass[letterpaper, 10 pt, conference]{ieeeconf}  
\usepackage{xcolor}

\IEEEoverridecommandlockouts                              

\usepackage{graphics} 
\usepackage{epsfig} 
\usepackage{mathptmx} 
\usepackage{times} 
\usepackage{amsmath} 
\usepackage{amssymb}  

\usepackage{comment}
\usepackage{epstopdf}
\usepackage{subfigure}
\usepackage{xspace}
\usepackage{multirow}
\usepackage{hyperref}
\usepackage{float}
\usepackage{balance}
\usepackage{siunitx,  
            booktabs, 
            caption}  
\usepackage{times}
\usepackage{soul}
\usepackage[utf8]{inputenc}
\usepackage{caption}
\definecolor{lightgray}{gray}{0.85}

\usepackage{amssymb} 
\usepackage{pifont}
\newif\ifcommenton
\commentontrue
\ifcommenton
\newcommand{\red}[1]{\textcolor{red}{#1}}
\newcommand{\blue}[1]{\textcolor{blue}{#1}}
\else
\newcommand{\red}[1]{}
\newcommand{\blue}[1]{}
\fi

\title{\LARGE \bf
Temporally-Continuous Probabilistic Prediction\\ using Polynomial Trajectory Parameterization
}

\author{ 
  Zhaoen Su, Chao Wang, Henggang Cui, Nemanja Djuric, Carlos Vallespi-Gonzalez, David Bradley \\
  \thanks{
   Uber Advanced Technologies Group (ATG), 50 33rd Street, Pittsburgh, PA 15201; emails:
     {\tt\small \{suzhaoen, chao.wang, hcui2, ndjuric, cvallespi, dbradley\}@uber.com}
}
}

\begin{document}

\maketitle
\thispagestyle{empty}
\pagestyle{empty}

\begin{abstract}
A commonly-used representation for motion prediction of actors is a sequence of waypoints (comprising positions and orientations) for each actor at discrete future time-points. While this approach is simple and flexible, it can exhibit unrealistic higher-order derivatives (such as acceleration) and approximation errors at intermediate time steps. To address this issue we propose a simple and general representation for temporally continuous probabilistic trajectory prediction that is based on polynomial trajectory parameterization. We evaluate the proposed representation on supervised trajectory prediction tasks using two large self-driving data sets. The results show realistic higher-order derivatives and better accuracy at interpolated time-points, as well as the benefits of the inferred noise distributions over the trajectories. Extensive experimental studies based on existing state-of-the-art models demonstrate the effectiveness of the proposed approach relative to other representations in predicting the future motions of vehicle, bicyclist, and pedestrian traffic actors.  
\end{abstract}

\section{Introduction}
In robotics in general and self-driving vehicle (SDV) applications in particular,  anticipating the motion of other actors around the robot plays a critical role in planning safe paths to navigate the environment~\cite{djuric2020}. Recently, significant improvements have come from exploring the input representation of the sensor data~\cite{chen2017multi, casas2018intentnet, zhou2019end, meyer2020laserflow, gao2020vectornet} and the neural network structures~\cite{luo2018fast, zhang2020stinet, djuric2020multixnet}. Likewise, the output representation for trajectories has seen extensions to account for multimodality~\cite{ivanovic2019trajectron, cui2019multimodal, zhao2020tnt} and for modeling probability distributions over their future locations~\cite{ djuric2020multixnet, alahi2016social}. However, these output representations generally provide predictions of locations only at discrete and prefixed time-points which may also lack the physics constraints that govern object motion in the real world. 

Prediction representations should offer enough flexibility to approximate the motion of various actors types, while still providing regularization that encourages physical realism in the predicted actor motion.  Physical realism, such as realistic velocities and accelerations in the trajectories, is particularly important for safety-critical applications.  Some representation choices provide such regularization~\cite{cui2020deep}, but also make learning more difficult by creating a more complex optimization surface. Additionally, prediction representations should be able to express multi-modal probability distributions over trajectories that reflect the uncertainty of the prediction. 
In robotics systems, trajectory predictions are often used to compare the probable future positions of other actors against possible future trajectories for the robot in order to find an efficient and low-risk trajectory for the robot~\cite{xu2014motion, raksincharoensak2016motion}. Many algorithms for this collision checking can be made more computationally efficient if actor locations can be accessed at arbitrary time-points in parallel within the prediction horizon. 


In this paper we propose an approach for trajectory prediction that exhibits these desirable properties. 
It expresses the time-varying distributions over actor future motion in terms of rigid transformations parameterized with polynomials. 
We show that low-order polynomials are effective for accurately representing labeled motion of various actor types.  When they are applied in extensive supervised learning tasks on two large-scale SDV data sets, the comparison of prediction performance shows that low-order polynomials are as effective as other representations at the fixed time-points, while providing better prediction accuracy on interpolated time-points, low-count actors, and learning tasks with large supervision time intervals. 
Moreover, we demonstrate that this representation implicitly provides effective regularization that improves physical realism in the predicted actor motion.



\section{Related work}
Commonly-used representations for trajectory predictions include waypoints and occupancy maps. The waypoint approach describes the probable future locations of an actor at some fixed, usually periodic, time-points~\cite{djuric2020}. In order to take the multimodality into account, multiple trajectories can be predicted for an actor~\cite{ivanovic2019trajectron, cui2019multimodal, zhao2020tnt}. When the uncertainty of the prediction is considered, a spatial probability distribution is provided at each of the given time-points independently~\cite{djuric2020multixnet,alahi2016social}. The mathematical details can also be found in the following section. The occupancy map representation expresses the multimodality and uncertainty of future actor motion by creating a spatially discretized grid around the actor. Each of the grid cells estimates the probability of the actor occupying this cell at a particular time-point~\cite{ kim2017probabilistic, xue2018ss, Oh_2020_CVPR}. 
Both representations, continuous in the spatial dimensions or not, are discrete temporally, which may lead to suboptimal performance in many real world applications where the actors usually behave smoothly. In this paper, we explore a prediction representation option that is continuous in both the spatial and temporal domains.

Physical realism of motion prediction is another important research topic. Studies in \cite{Cui2020LearningDR, li2020dynamic} improve the feasibility of human motion prediction by constructing the graph of skeleton joints and applying constraints on the graph edges. 
Physical realism such as collision avoidance and kinematic feasibility is studied for vehicle trajectory prediction. 
In the interactive vehicle following scenario, physical models are embedded in a network to avoid collision~\cite{tang2019adaptive}. 
The authors in~\cite{cui2020deep} build a vehicle kinematic model for the waypoint representation with constraints and regularization to enforce kinematic feasibility. 
We show that the proposed low-order polynomial representation leads to good physical realism in predicted trajectories, without enforcing additional physical models, constraints, or regularization.

Lastly, it is well-known that low-dimensional parametric approximation is widely used in many scientific and engineering fields, such as the spectrum approximation based on Gaussian, Lorentzian, or Voigt functions in spectroscopy~\cite{whiting1968empirical,su2018prl, su2020prb}, or the value function approximation in reinforcement learning~\cite{baird1995residual,gordon1995stable, sutton2000policy}. 
Its benefits such as providing implicit regularization, compact expression, and avoiding the curse of dimensionality, are also well-understood. In this work, we successfully apply this methodology to the task of motion prediction. 



\section{Methodology}
\subsection{Trajectory representation}
The waypoint trajectory representation $\mathcal{P}$ can be generally expressed as a sequence of rigid transformations, SE3 in general or SE2 for 2D applications,

\begin{equation}
    \label{eq:wpt}
    \mathcal{P} = \{({\bf{T}}_t, {\bf{R}}_t)\}, t \in \{0, t_1, t_2, \ldots, T\},
\end{equation}
each of which denotes the translation and rotation of the actor at time $t$.  
The probabilistic representation describes the full probability distribution in terms of a sequence of $(p({\bf{T}}_t), p({\bf{R}}_t))$ each of which denotes the spatial probability densities of the position and orientation at time $t$. 
A probability distribution can be described with its sufficient statistics in terms of moments or distribution parameters if it has an analytical expression, denoted as $\bf{M}$. 
Note that the non-probabilistic representation in (\ref{eq:wpt}) can be viewed as a special case where only the zeroth moments (i.e., the means) are considered. 
While the waypoint representation is very flexible to express an arbitrary trajectory at the predicted time-points, intermediate time-points need to be interpolated, with a common choice being linear interpolation (i.e., the trajectory is interpreted as a linear spline). As linear interpolation introduces approximation errors on accelerating objects, this might be mitigated by predicting many time-points which increases the computational complexity.

We propose another general prediction representation which parameterizes the prediction distributions over time based on polynomial approximation. 
More specifically, we represent each scalar element $m$ of $\bf{M}$ independently with a low-order polynomial function of time as follows,

\begin{equation}
    \label{eq:moment}
    m(t) = f_m\left(\sum_{n=0}^{N_m}{a_{m,n} \left(\frac{t}{T}\right)^n}\right),
\end{equation}
where $a_{m,n}$ are the coefficients of the polynomial of degree $N_m$, and $f_m(\cdot)$ can be an identity function or a function to ensure the validity of $\bf{M}$, if necessary. The normalization over the maximum prediction horizon of interest $T$, is often desired in practice, particularly for large $t$'s. 
In this paper, we explore the trajectory prediction of traffic actors in SDV applications where each actor can be approximated with a fixed polygon and a time-varying SE2 transformation from the frame of the polygon to a shared \emph{world} frame.  The SE2 transformation consists of translation in the $x$-$y$ plane and yaw rotation around the vertical axis, denoted as $({c_{x}}_t, {c_{y}}_t, {\sin\theta}_t, {\cos\theta}_t)$. We model each of the components independently with a univariate distribution. Using the Laplace distribution as an example, the probabilistic prediction for a specific component $v$ can be expressed as
\begin{equation}
    \label{eq:prob}
    P_v(t) = \mathcal{L}(v|\mu_{v}(t), b_{v}(t)),
\end{equation}
whose means over time $t$ are parameterized by a polynomial 

\begin{equation}
    \label{eq:mean}
    \mu_{v}(t) = \sum_{n=0}^{N_{\mu_{v}}}{a_{\mu_{v},n} \left(\frac{t}{T}\right)^n}.
\end{equation}
One physics insight into the coefficients can be found by noticing that the important moments of the trajectory can be expressed analytically and computed from the coefficients at arbitrary time-points (without finite differencing), including position, velocity, acceleration, lateral acceleration, curvature, etc., for $N_{\mu_{v}} > 1$. The waypoint representation with linear interpolation, in contrast, is equivalent to using polynomials with $N_{\mu_{v}} = 1$ over each time interval, and it only allows for direct computation of position. Another view is that predicting the polynomial coefficients is equivalent to predicting $N_{\mu_{v}}+1$ control points that determine a trajectory through polynomial interpolation. Besides $\mu_{v}$, the diversity parameter $b_{v}$ over time can be parameterized as

\begin{equation}
    \label{eq:unc}
    b_{v}(t) = \exp{\sum_{n=0}^{N_{b_{v}}}{a_{b_{v},n} \left(\frac{t}{T}\right)^n}}.
\end{equation}
The additional exponential function ensures the positiveness. Note that the generality and simplicity of the representation allow its application in common non-probabilistic and probabilistic prediction models of various spatial distributions, and straightforward replacement of the representation in the models that output waypoints with few changes.

\subsection{Label trajectory approximation}

\begin{figure*}[t!]
\vspace{0.1cm}
    \centering
    \includegraphics[width=0.295\textwidth]{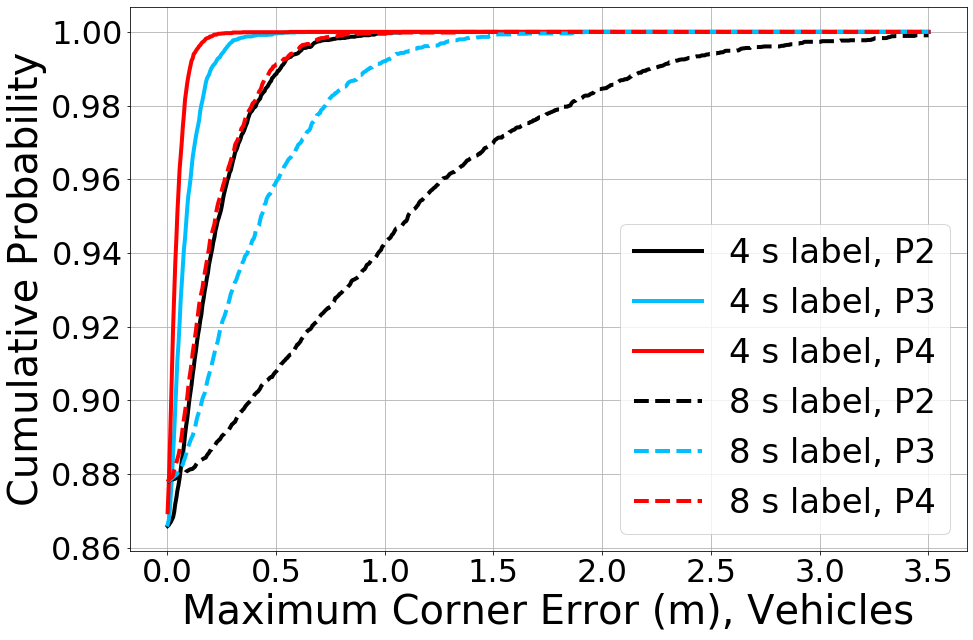}
    \includegraphics[width=0.285\textwidth]{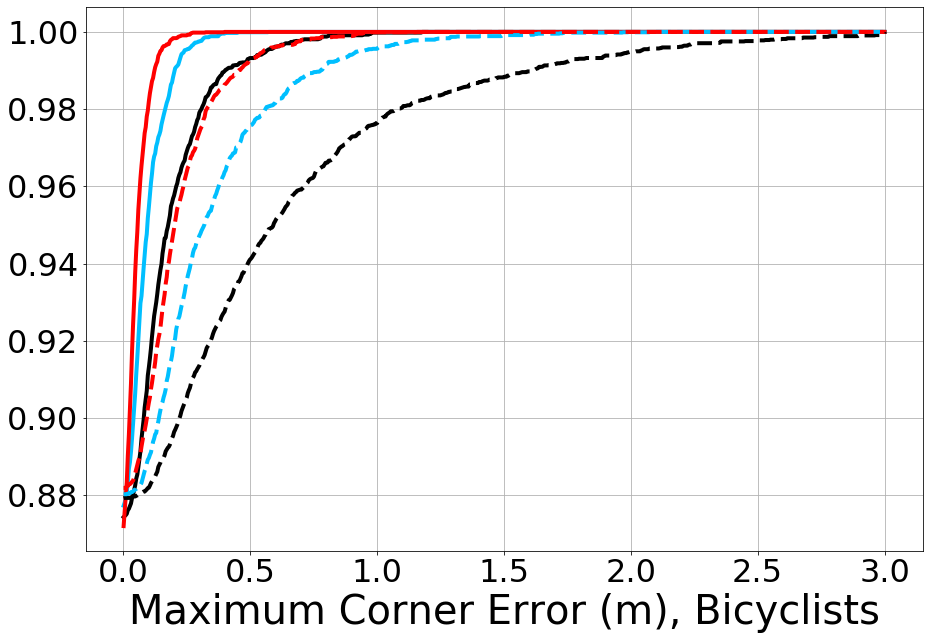}
    \includegraphics[width=0.28\textwidth]{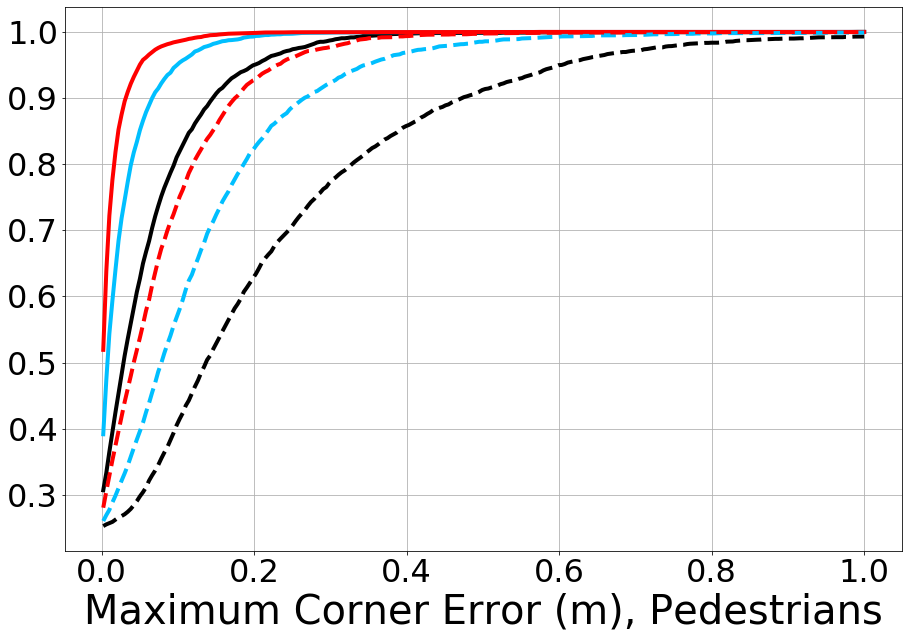}
    \caption{Approximation errors of polynomial representation to fit label trajectories. Cumulative fraction as a function of max corner errors for $4$s and $8$s trajectories of vehicles, bicyclists, and pedestrians, with polynomials of degrees 2-4 (P$2$-$4$)}
    \label{fig:label_fitting}
\vspace{-0.2cm}
\end{figure*}

The Weierstrass Approximation Theorem~\cite{weierstrass1988} states that any continuous function can be approximated to arbitrary accuracy on a closed and bounded interval with a polynomial of sufficiently high degree. On the in-house data set (see its details in the next section), we investigate the maximum approximation error using (\ref{eq:mean}) to fit label trajectories with various low-order polynomials and prediction horizons. The label trajectory of each actor in the data set can be represented by a bounding box of fixed size with a sequence of SE2 transformations at $10$Hz. We parameterize the four components ($c_x$, $c_y$, $\sin\theta$, $\cos\theta$) of the time-varying SE2 transform as follows, using two polynomials for the centroid translation (i.e., $c_x$ and $c_y$) and two polynomials which are normalized to produce the sine and cosine components of the rotation. We find the best-fit polynomials by minimizing the total L2 error between the labeled bounding box corner and the associated fitted corner over all corners and all 4-second time-points at $10$Hz. 
While for metrics, we compute the maximum corner error for each trajectory defined as the maximum L2 distance over all corners and all time-points, which examines the worst approximation.  Note this error metric includes both translation and orientation errors, and bounds the maximum error that a collision-checking algorithm could encounter.  

Fig.~\ref{fig:label_fitting} shows the maximum corner error using polynomials of degrees $2$ through $4$ to fit label trajectories over $4$s and $8$s time horizons, for vehicles, bicyclists, and pedestrians. As expected, polynomials of higher orders yield lower approximation errors. However, what is surprising is how quickly the maximum corner error drops below the average predicted centroid displacement error as the order of the polynomial is increased, suggesting that even very low-order polynomials are able to represent traffic actor motion with small approximation error relative to the expected prediction error of state-of-the-art prediction models. For instance, the average displacement error for vehicles at $8$s is approximately $1.3$m (see Table~\ref{tab:metrics_8s}), and $96.0\%$ of quadratic trajectories and $99.5\%$ of cubic trajectories have maximum corner error less than that value.  Further analysis shows that the high approximation errors occur mainly on maneuvers that have high jerk, or switch between being stopped and moving. For models that cannot achieve good predictions for such hard maneuvers, the low-order polynomials would not hurt; higher-order polynomials or splines of polynomials can be used to provide more representational capacity when necessary. This paper focuses on low-order polynomials as the results in Table~\ref{tab:metrics_8s} show no significant gains in prediction accuracy between $3$rd and $4$th order polynomials for prediction horizons of up to $8$s.  

\subsection{Applying the representation in supervised learning}
Next, we study the proposed representation in supervised trajectory prediction tasks by replacing the waypoint representation with the polynomial representation using (\ref{eq:prob}), and compare prediction performances using the different representations. We adapt MultiXNet\cite{djuric2020multixnet}, which is a deep model with competitive performance designed to detect traffic actors around a SDV and predict their future trajectories. Like most works in trajectory prediction~\cite{lee2017desire, ivanovic2019trajectron, salzmann2020trajectron++, gupta2018social, zhao2019multi, sadeghian2019sophie, kosaraju2019social}, MultiXNet outputs trajectory prediction at periodic time-points $t \in \{0, \tau, 2\tau, \ldots, n\tau\}$. The model assumes a univariate Laplace distribution for each of the ($c_x$, $c_y$, $\sin\theta$, $\cos\theta$) components at each of the $n+1$ time-points, i.e, $(n+1)$ $\times$ $4$ $\times$ $2$ values are regressed for the means ($\mu$'s) and diversity parameters ($b$'s). By contrast, in models with the polynomial representation, the regression values are the $N_{\mu}+1$ coefficients in (\ref{eq:mean})  and the $N_{b}+1$ coefficients in (\ref{eq:unc}), for the four components individually. If multimodal prediction is modeled, such as for the vehicles in MultiXNet, independent polynomial representation is applied to each separate mode. 
 
Note that the data sets used in this paper provide prediction label as bounding boxes at periodic time-points. To train the polynomial models, we sample waypoints from the polynomial prediction at the same time-points as the waypoint models, and the regression loss is applied to these sampled waypoints. While it would be possible to compute the target polynomial coefficients by fitting the label trajectories and define regression loss on the polynomial coefficients directly, our goal in this study is to compare representations with minimal modeling differences, therefore we keep the regression targets and loss functions unchanged. 

\section{Evaluation}

\begin{table*} [!htbp]
\vspace{0.1cm}
\centering
\small 
\caption{Four-second models on nuScenes and the in-house data sets with different representations for the means. First block: model comparison on nuScenes data set with a supervision interval of $0.5$s. The models using waypoints and polynomials of degrees $1$-$3$ for the means are denoted as WP and P$1$-$3$. Second block: model comparison on the in-house data set with a supervision interval of $0.1$s. On top of WP models, KM is only applied to vehicle prediction. DE is in meters, and $\Delta\theta$ is in degrees.  Lowest errors within metric variance are in bold.}
\label{tab:metrics_4s}
\vspace{0.2cm}
  \begin{tabular}{lcccccccccccccc}
    & \multicolumn{4}{c}{\bf Vehicles} & \multicolumn{4}{c}{\bf Bicyclists} & \multicolumn{2}{c}{\bf Pedestrians} \\
    \cmidrule(lr){2-5} \cmidrule(lr){6-9} \cmidrule(lr){10-11}
    {\bf Method}  & {\bf 2s DE} & {\bf 4s DE} & {\bf 2s $\Delta\theta$} & {\bf 4s $\Delta\theta$} &  {\bf 2s DE} & {\bf 4s DE} & {\bf 2s $\Delta\theta$} & {\bf 4s $\Delta\theta$}  & {\bf 2s DE} & {\bf 4s DE} \\
    \hline
    WP & 	{\bf 0.73}&	1.67&	{\bf 2.39}&	{\bf 3.33}&		{\bf 1.6}&	3.6&	7.7&	{\bf 10.5}&		{\bf 0.51}&	1.06 \\
    P1 & 	0.75&	1.74&	2.50&	3.49&		{\bf 1.5}&	{\bf 3.1}&	8.0&	10.7&		{\bf 0.51}&	{\bf 1.05} \\
    P2 & 	{\bf 0.73}&	1.68&	{\bf 2.37}&	{\bf 3.39}&		{\bf 1.6}&	{\bf 3.4}&	{\bf 7.4}&	{\bf 10.3}&		{\bf 0.51}&	1.06 \\
    P3 & 	{\bf 0.74}&	{\bf 1.65}&	{\bf 2.41}&	{\bf 3.41}&		1.9&	3.6&	{\bf 7.6}&	{\bf 10.6}&		{\bf 0.50}&	{\bf 1.04} \\
    \hline  \hline
    WP & 	{\bf 0.307}&	{\bf 0.565}&	{\bf 1.47}&	{\bf 1.76}&		{\bf 0.27}&	0.52&	{\bf 5.8}&	{\bf 6.0}&		{\bf 0.378}&	{\bf 0.807} \\
    P1 & 	0.328&	0.672&	1.51&	1.83&		{\bf 0.27}&	{\bf 0.51}&	6.2&	6.4&		{\bf 0.380}&	0.815 \\
    P2 & 	0.311&	{\bf 0.563}&	{\bf 1.45}&	{\bf 1.79}&	{\bf 0.28}&	{\bf 0.50}&	{\bf 5.9}&	{\bf 6.1}&	{\bf 0.379}&	{\bf 0.811} \\
    P3 & 	0.311&	0.568&	1.49&	{\bf 1.79}&		{\bf 0.28}&	{\bf 0.51}&	6.1&	6.4&		{\bf 0.380}&	0.812 \\
    KM & 	0.313&	0.575&	1.64&	1.80&		0.30&	0.53&	{\bf 5.9}&	{\bf 6.1}&		{\bf 0.381}&	0.812 \\
    \hline
\end{tabular}
\vspace{-0.1cm}
\end{table*}

\begin{table*} [!htbp]
\centering
\small 
\caption{Eight-second models on the in-house data with a supervision interval of $0.1$s. The models using waypoints and polynomials of degrees $1$-$4$ for the means are denoted as WP and P$1$-$4$.} 
\label{tab:metrics_8s}
\vspace{0.2cm}
  \begin{tabular}{lcccccccccccccc}
    & \multicolumn{4}{c}{\bf Vehicles} & \multicolumn{4}{c}{\bf Bicyclists} & \multicolumn{2}{c}{\bf Pedestrians} \\
    \cmidrule(lr){2-5} \cmidrule(lr){6-9} \cmidrule(lr){10-11}
    {\bf Method} &  {\bf 4s DE} & {\bf 8s DE} & {\bf 4s $\Delta\theta$} & {\bf 8s $\Delta\theta$} &  {\bf 4s DE} & {\bf 8s DE} & {\bf 4s $\Delta\theta$} & {\bf 8s $\Delta\theta$} &  {\bf 4s DE} & {\bf 8s DE} \\
    \hline
    WP& 	{\bf 0.580}&	1.362&	{\bf 1.78}&	{\bf 2.21}&		0.70&	1.41&	{\bf 6.5}&	6.8&	{\bf 0.828}&	{\bf 1.903} \\
    P1& 	0.684&	1.618&	1.85&	2.36&		0.67&	1.28&	6.7&	7.0&		0.832&	1.926 \\
    P2&		0.593&	1.295&	1.83&	2.31&	{\bf 0.58}&	{\bf 1.13}&	6.7&	6.9&	{\bf 0.826}&	{\bf 1.899} \\
    P3&		0.590&	{\bf 1.291}&	1.82&	2.28&	{\bf 0.59}&	1.21&	{\bf 6.5}&	{\bf 6.7}&		{\bf 0.827}&	{\bf 1.899} \\
    P4&		0.595&	{\bf 1.287}&	1.82&	2.28&	{\bf 0.60}&	1.26&	{\bf 6.4}&	{\bf 6.6}&		{\bf 0.829}&	1.913 \\
    \hline    
\end{tabular}
\vspace{-0.1cm}
\end{table*}

\begin{table*} [!htbp]
\centering
\small 
\caption{Comparison of using waypoints (WP) and polynomials of degree $2$ (P$2$) on simplified variants that are single-stage and single-modal, and uses displacement errors as the regression losses. The experiments are performed on the in-house data set and have four-second predictions with interval $0.1$s.} 
\label{tab:metrics_vanilla}
\vspace{0.2cm}
  \begin{tabular}{lcccccccccccccc}
    & \multicolumn{4}{c}{\bf Vehicles} & \multicolumn{4}{c}{\bf Bicyclists} & \multicolumn{2}{c}{\bf Pedestrians} \\
    \cmidrule(lr){2-5} \cmidrule(lr){6-9} \cmidrule(lr){10-11}
    {\bf Method}  & {\bf 2s DE} & {\bf 4s DE} & {\bf 2s $\Delta\theta$} & {\bf 4s $\Delta\theta$} &  {\bf 2s DE} & {\bf 4s DE} & {\bf 2s $\Delta\theta$} & {\bf 4s $\Delta\theta$}  & {\bf 2s DE} & {\bf 4s DE} \\
    \hline
    WP 	&{\bf 0.337}	&{\bf 0.653}	&{\bf 1.68}	&{\bf 2.05}	&{\bf 0.29}	&{\bf 0.52}	&{\bf 6.5}	&{\bf 6.8}	&{\bf 0.414}	&0.874 \\
    P2	&{\bf 0.339}	&{\bf 0.654}	&{\bf 1.68}	&{\bf 2.04}	&{\bf 0.28}	&{\bf 0.51}	&{\bf 6.6}	&{\bf 6.9}	&{\bf 0.411}	&{\bf 0.870} \\    
    \hline
\end{tabular}
\vspace{-0.1cm}
\end{table*}

\subsection{Experimental setups}
{\bf Implementation details.}
We evaluate the proposed representation by adapting the MultiXNet as the waypoint representation (WP) baseline.
In each comparison experiment group, the supervision is provided at the same periodic time-points. The baseline WP model outputs waypoint probabilistic prediction in terms of four univariate Laplace distributions for $({c_{x}}_t, {c_{y}}_t, {\sin\theta}_t, {\cos\theta}_t)$ at each of the time-points (note however that we do not model heading for pedestrians). To simplify the discussion, we use polynomial representation of a same degree $d$ for all of the four means, denoted as P$d$, and polynomial of a same degree $k$ for their diversity parameters, denoted as P$k$($b$). Using the 8-second prediction as an example, a waypoint model would regress 648 values for one trajectory, while the polynomial representation for ($d = 2$, $k = 1$) would provide 20 values as the model output instead. We use this setting as a default for the polynomial models, unless specified differently. We also implement the vehicle kinematic model (KM) proposed in~\cite{cui2020deep, kong2015kinematic}. The KM model outputs the longitudinal acceleration and curvature for each waypoint, clipped within $[-8, 8]~\textrm{m/s}^2$ and $[-0.2, 0.2]~\textrm{m}^{-1}$, respectively, with L2 regularization with weight $0.1$ applied on the outputs to encourage smoothness.

{\bf Data.}
The experiments are carried out on the open-sourced nuScenes data~\cite{caesar2020nuscenes} ($7000$ scenes of $20$s in the training split with $2$Hz annotations) with the results shown in Table~\ref{tab:metrics_4s}. We also used the larger Uber 
ATG in-house data set ($14000$ scenes of $25$s in the training split with $10$Hz annotations) throughout the studies, as it yields lower metric variances and the finer label interval facilitates finite difference computation. We explore short-term ($4$s) and mid-term ($8$s) prediction for (a) vehicles that are the most common traffic actors, (b) bicyclists that are as rare as about $2\%$ of vehicles in the data sets, and (c) pedestrians whose motion can change abruptly.

{\bf Metrics.}
Detection is evaluated using average-precision (AP) with the intersection-over-union (IoU) matching threshold of $0.5$, $0.3$, $0.1$ for vehicles, bicyclists, and pedestrians, respectively. We report prediction performance in terms of displacement errors (DE's) for centroids, and angle error ($\Delta{\theta}$'s) for headings. The models within each comparison group have close AP's and are thus not reported. To compare prediction fairly, the prediction metrics are computed with the detection probability threshold set to yield a recall of $0.8$ as the operational point for the models trained on the in-house data set, and $0.6$, $0.3$, and $0.6$ for vehicles, bicyclists, and pedestrians, respectively on the nuScenes data set. 

\subsection{Prediction performance}
\begin{figure*}[!htbp]
\vspace{0.1cm}
    \centering
    \includegraphics[width=0.234\textwidth]{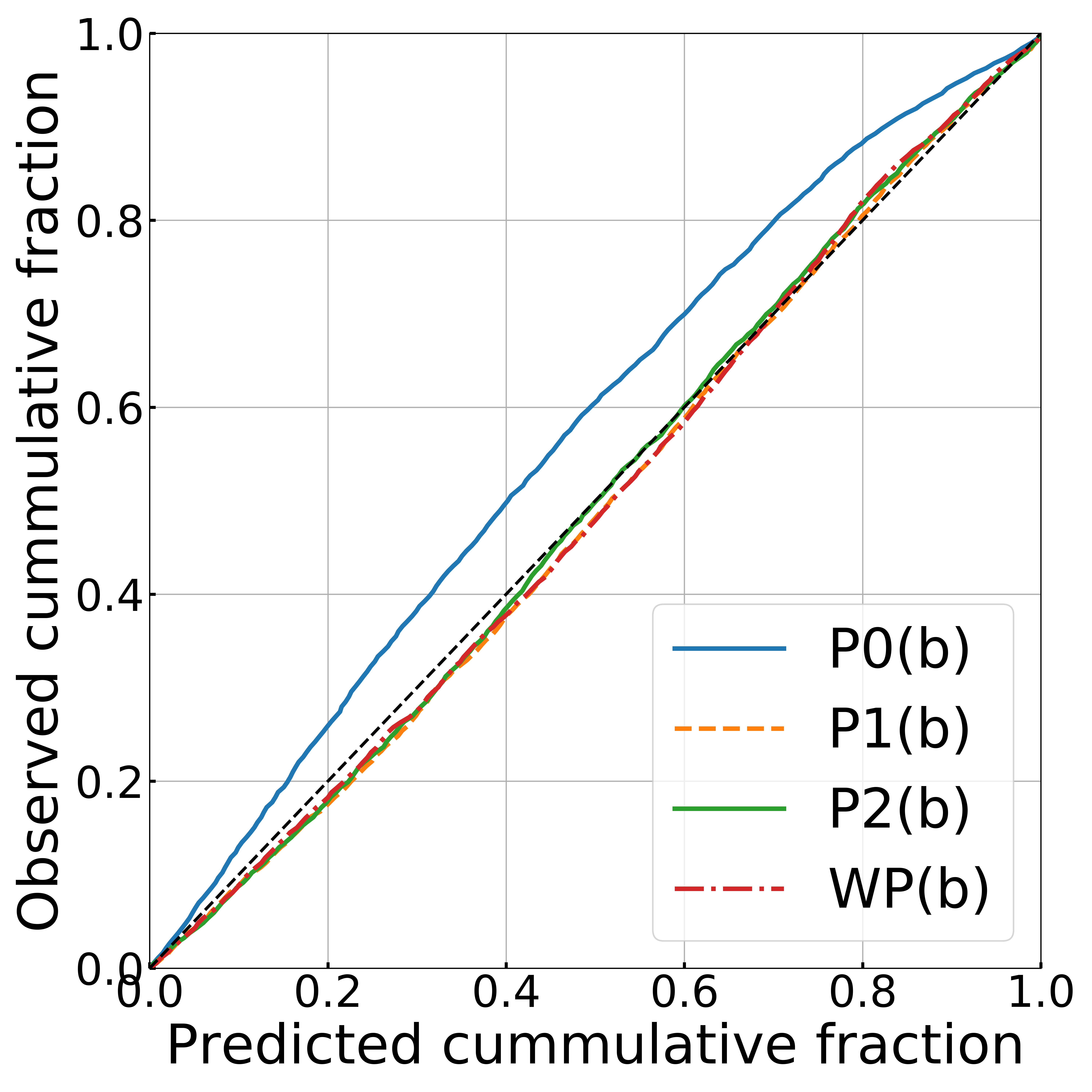}
    \includegraphics[width=0.23\textwidth]{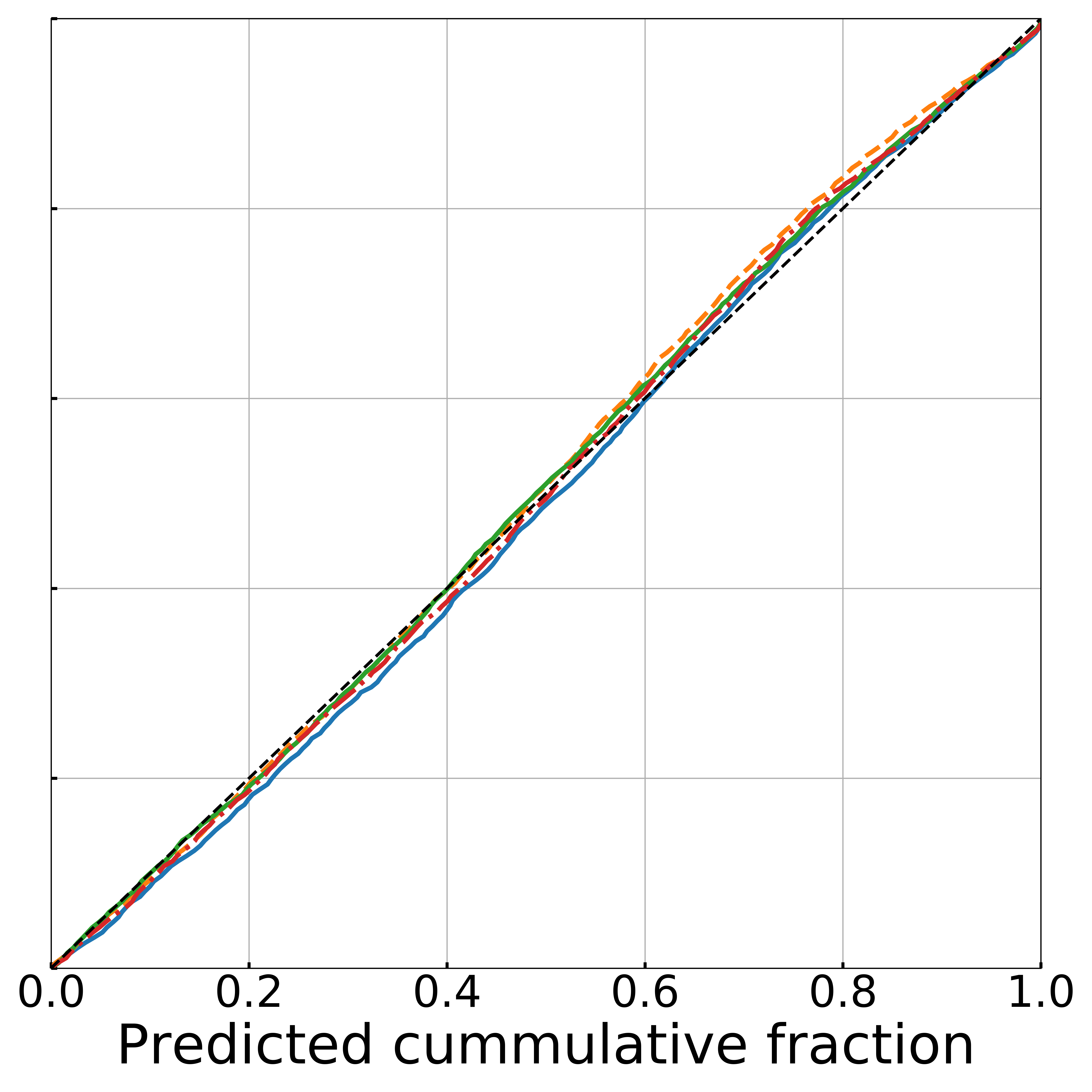}
    \includegraphics[width=0.23\textwidth]{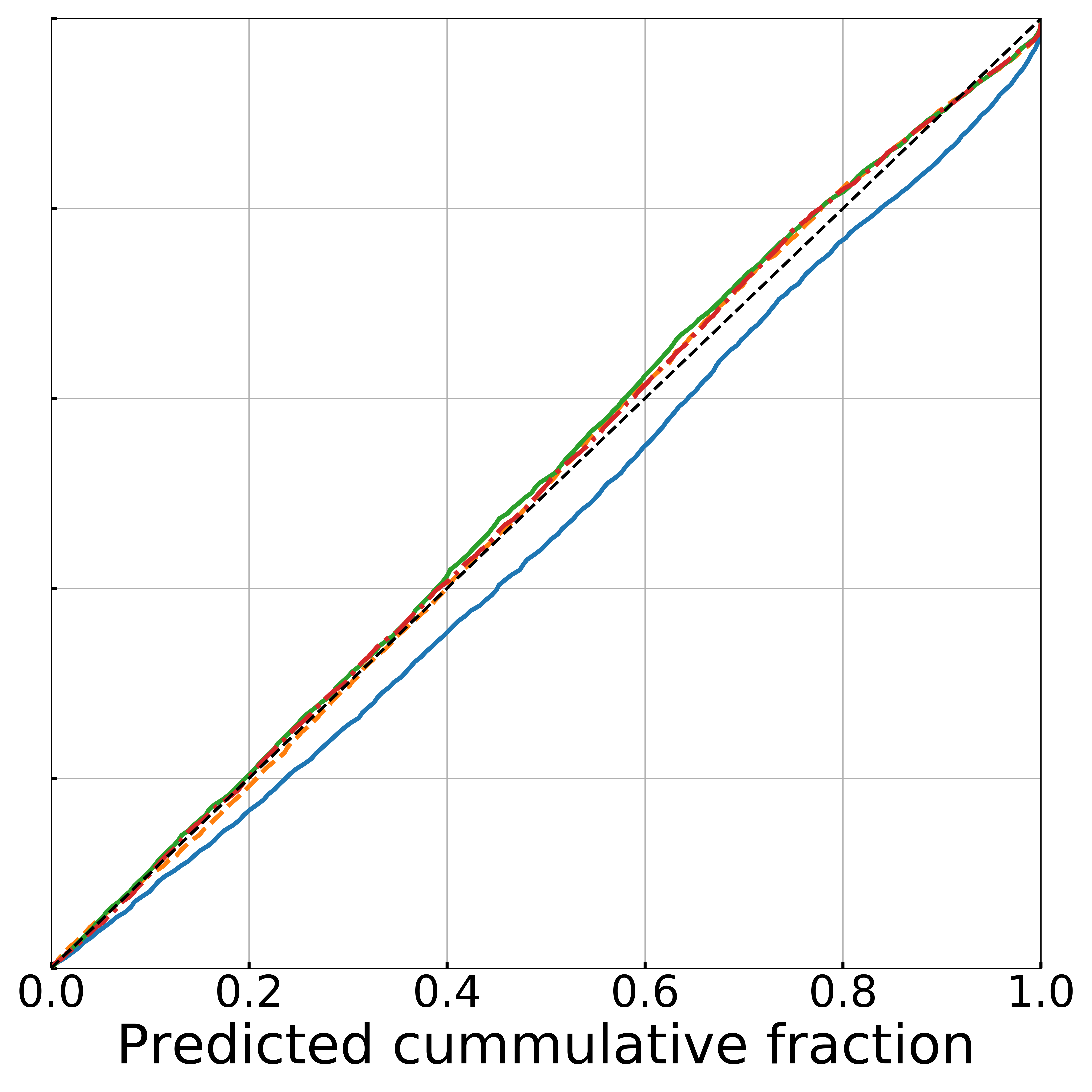}
    \caption{Reliability diagrams for the cross-track dimension at $0$s (left), $2$s (middle) and $4$s (right) of models using waypoints (WP($b$)) and polynomials of degrees $0$-$2$ (P$0$-$2$($b$)) for the diversity parameters.}
    \label{fig:RD}
    \vspace{-0.1cm}
\end{figure*}

\begin{table*} [!htbp]
\centering
\small 
\caption{Comparison of the continuous prediction using polynomial of degree $2$ (P$2$) and the discrete prediction using waypoints (WP) for the means on the in-house data set. The regression supervision is provided at $0$, $2$, and $4$s. DE in meters is provided in the first block. $\Delta\theta$ in degrees is in the second block. The predictions at $1$ and $3$s of WP are computed by linear interpolation. Metrics are computed only for actors that are faster than $0.2$m/s. 
}
\vspace{0.2cm}
\label{tab:metrics_inter}
  \begin{tabular}{lcccccccccccccc}
    & \multicolumn{4}{c}{\bf Vehicles} & \multicolumn{4}{c}{\bf Bicyclists} & \multicolumn{4}{c}{\bf Pedestrians} \\
    \cmidrule(lr){2-5} \cmidrule(lr){6-9} \cmidrule(lr){9-13}
    {\bf Method}  & {\bf 1s DE} & {\bf 2s DE} & {\bf 3s DE} & {\bf 4s DE}  & {\bf 1s DE} & {\bf 2s DE} & {\bf 3s DE} & {\bf 4s DE} & {\bf 1s DE} & {\bf 2s DE} & {\bf 3s DE} & {\bf 4s DE}\\
    \hline
    WP 	&0.99	&{\bf 1.92}	&3.43	&4.97	&3.9	&7.7	&11.1	&14.2	&{\bf 0.34}	&{\bf 0.63} &{\bf 0.96}	&{\bf 1.30}\\
    P2 	&{\bf 0.92}	&{\bf 1.90}	&{\bf 3.21}	&{\bf 4.85}	&{\bf 2.9}	&{\bf 5.1}	&{\bf 6.8}	&{\bf 8.3}	&{\bf 0.34}	&{\bf 0.64}  &{\bf 0.95}	&{\bf 1.29}\\
    \hline
    {} & {\bf 1s $\Delta\theta$} & {\bf 2s $\Delta\theta$} & {\bf 3s $\Delta\theta$} & {\bf 4s $\Delta\theta$} & {\bf 1s $\Delta\theta$} & {\bf 2s $\Delta\theta$} & {\bf 3s $\Delta\theta$} & {\bf 4s $\Delta\theta$} & {\bf 1s $\Delta\theta$} & {\bf 2s $\Delta\theta$} & {\bf 3s $\Delta\theta$} & {\bf 4s $\Delta\theta$} \\    
    \hline
    WP &1.97	&2.92	&4.13	&{\bf 5.59}	&{\bf 4.6}	&6.6	&8.2	&9.0	& - & - & - & -	\\
    P2 &{\bf 1.87}	&{\bf 2.81}	&{\bf 4.06}	&{\bf 5.56}	&{\bf 4.6}	&{\bf 6.4}	&{\bf 7.9}	&{\bf 8.7} & - & - & - & -		\\
    \hline
\end{tabular}
\vspace{-0.2cm}
\end{table*}

The first block in Table~\ref{tab:metrics_4s} shows the experiments on nuScenes data with prediction supervision at $t \in \{0, 0.5, 1.0, \ldots, 4.0\}$ seconds. Within the metric variance, P$2$ and P$3$ achieve performance similar to WP. P$1$ has worse performance for vehicles and pedestrians, which we attribute to its lack of representational power. P$1$ outperforms other models for predicting bicycle centroids, which might be explained by the low population of bicyclists in the data set and the strong regularization provided by  polynomials of degree $1$. 
Similarly $4$-second prediction is studied on the in-house data set with interval $0.1$s, and presented in the second block. With lower metric variance, we again see that the performance of P$2$ and P$3$ is close to that of WP for vehicles and pedestrians. The under-performance of P$1$ for vehicles and pedestrians is further confirmed on this data set. Furthermore, the in-house data set uses a finer supervision interval which might explain why P$1$ does not outperform the other models on the  bicyclists class. Then, we further extend the prediction supervision horizon to $8$s (See Table~\ref{tab:metrics_8s}). Except for P$1$, which is too simple to express $8$s trajectories, the polynomial models still perform as well as the model using waypoints. 
To show that the representation works effectively with other model designs and loss functions, we study the representations in single-stage single-modal MultiXNet variants where the second stage network is removed, each actor is modeled with a single trajectory, and the Kullback–Leibler divergence for trajectory regression is replaced by the displacement error without uncertainty learning using the smooth-L1 loss~\cite{djuric2020}. For the models with these common but less optimal techniques, the prediction accuracy is also close using the polynomials and waypoints (see Table~\ref{tab:metrics_vanilla}).


We measure the calibration of the probabilistic predictions using waypoint representation (WP($b$)), and polynomials of degrees $0$-$2$ (P$0$-$2$($b$)) for the diversity parameters. Fig.~\ref{fig:RD} provides one example by their reliability diagrams \cite{djuric2020} for the cross-track dimension at $0$s, $2$s, and $4$s. Except for P$0$($b$), the probabilistic predictions are well calibrated in all models, as their curves are close to the reference lines. P$0$($b$) has stationary diversity parameters, which yields under confident predictions at the start of the prediction horizon, and over-confident prediction at the end of the horizon. Notice that polynomials of degree $1$ suffice for representing $b$ in these $4$-second models.

\subsection{Continuous prediction representation}
We hypothesize that the polynomial representation improves accuracy over linear interpolation at time-points where supervision is not provided during training.
To demonstrate this, we study models with waypoint representations (WP) and polynomials of degree $2$ (P$2$) for the means with supervision at $0$s, $2$s, and $4$s. We compare their performance at $1$s, $2$s, $3$s, and $4$s. The predictions of WP at $1$s and $3$s are computed by linear interpolation. Table~\ref{tab:metrics_inter} focuses on the actors faster than $0.2$m/s. For the prediction of vehicles, P$2$ has slightly better performance at $2$s and $4$s where regression supervision is available, while it outperforms marginally the interpolated prediction of WP at $1$s and $3$s. WP has significantly worse performance in centroid prediction of bicyclists at all time-points, which can be explained by the large supervision interval and low population in the training, while the regularization provided by low-order polynomials mitigates those problems. Notice that for predictions of vehicles and bicyclists, the polynomial representation exhibits strength over the waypoint representation even at the fixed time-points, when the supervision time intervals are large, in additional to the better accuracy at intermediate time-points. The two models perform similarly for pedestrians, suggesting that the second order states for pedestrians are either not captured by the models or less important in pedestrian prediction.

\begin{figure*}[!htbp]
\vspace{0.1cm}
    \centering
    \includegraphics[width=0.349\textwidth]{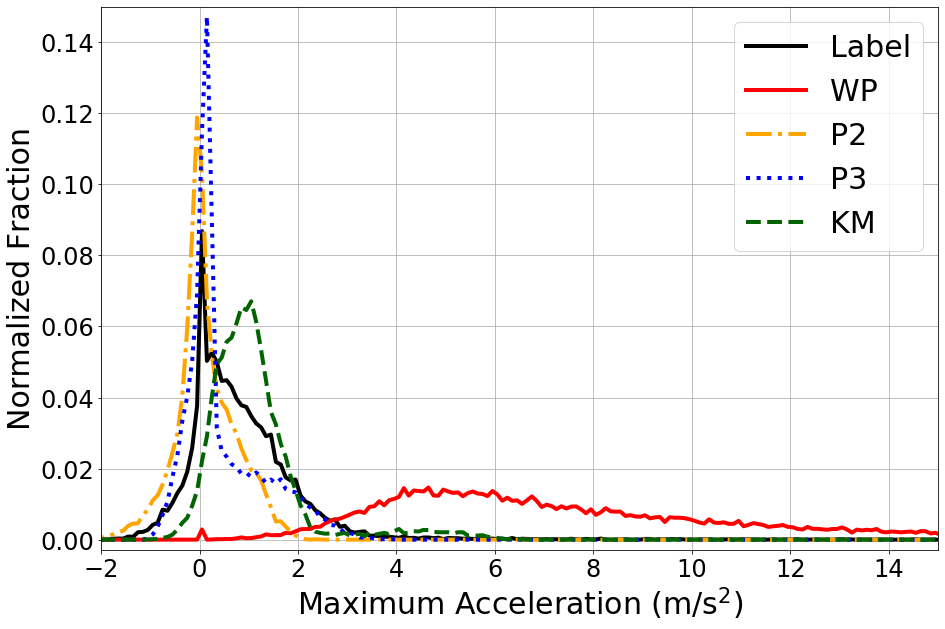}
    \includegraphics[width=0.339\textwidth]{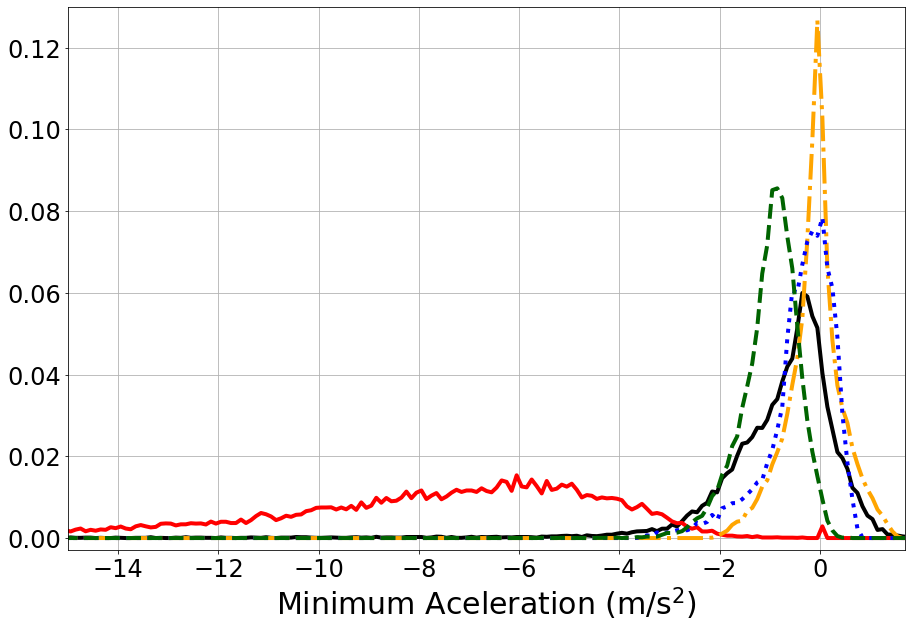}
    \includegraphics[width=0.35\textwidth]{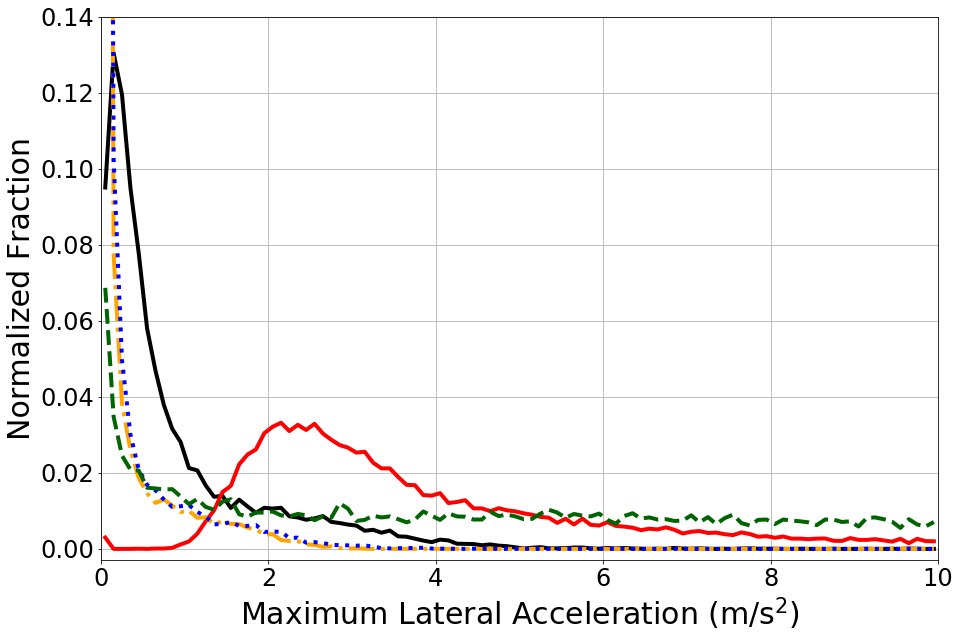}
    \includegraphics[width=0.35\textwidth]{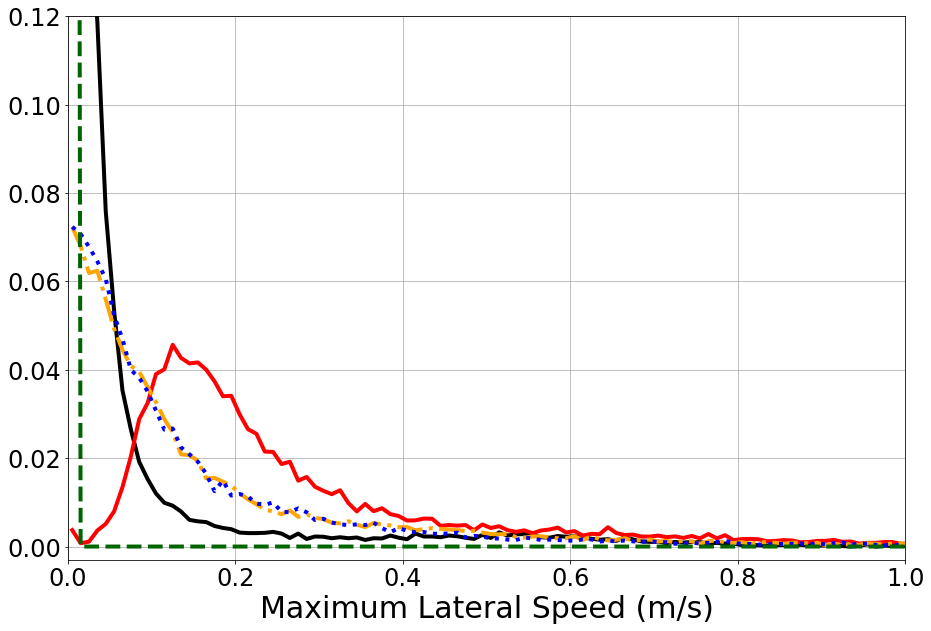}
    \caption{Normalized histograms of maximum acceleration, minimum acceleration, maximum lateral acceleration, and maximum lateral speed of label trajectories (Label) and prediction trajectories by the models (WP, P$2$, P$3$ and KM). Y-axis values of the plots are fractions of non-static vehicle trajectories per $0.1$, $0.1$, $0.1$ $\textrm{m/s}^2$, and $0.01$ m/s interval, respectively. The label and all models have $4$-second trajectories. The prediction trajectories of P$2$-$3$ are sampled with the same $0.1$s interval for the metric computation that uses finite difference.}
    \label{fig:hist_feasibility}
\vspace{-0.4cm}
\end{figure*}

\subsection{Physical feasibility}
To measure the physical realism of a trajectory, we analyze the maximum and the minimum longitudinal acceleration (${\bf a}_t \cdot {\bf v}_t /|{\bf v}_t|$), maximum lateral acceleration (${\bf a}_t \times {\bf v}_t /|{\bf v}_t|$), and maximum lateral speed  ($|{\bf v}_t \times {\bf h}_t|$) over all prediction time-points for each trajectory, where ${\bf a}_t$ is the centroid acceleration vector, ${\bf v}_t$ is the centroid velocity vector, and ${\bf h}_t  = (\cos\theta_t, \sin\theta_t)$ is the unit heading vector, measured at time-point $t$. For motion of common traffic actors these quantities are usually tightly constrained, as shown in Fig.~\ref{fig:hist_feasibility} by their distribution for the label trajectories on the in-house data set. We compare this label distribution to the trajectory distributions predicted by models using different representations.

Fig.~\ref{fig:hist_feasibility} shows their distributions in normalized histograms that focus on trajectories of non-static vehicles. One can see that the majority of the trajectories of WP have infeasible maximum and minimum accelerations, while the trajectories of label and P$2$-$3$ have distributions concentrating near zero. Interestingly, KM trajectories peak at about $\pm 1\textrm{m/s}^2$ instead of close to zero.
In the maximum lateral acceleration plot, a large portion of the WP trajectories have less feasible lateral acceleration, while trajectories of the polynomial models demonstrate closer distributions to that of labels. Because there are no constraints on lateral acceleration in KM, it has considerable amount of trajectories showing unfeasible lateral acceleration. Lastly, the trajectories of WP show the greatest divergence from the label distribution in lateral speed. All trajectories produced by KM have zero lateral speed as it is enforced by the explicit vehicle model. Note that non-zero lateral speed is expected in the label trajectories and those of P$2$-$3$ because the speed is computed for the centroids instead of the rear axle centers that are not annotated. Feasible longitudinal acceleration and lateral speed in KM are achieved by hand-crafted constraints and regularization on the controls, which may contribute to unnatural behaviors such as the distribution peaks at around $1\textrm{m/s}^2$ and $5\textrm{m/s}^2$ in the maximum longitudinal distribution, and the prediction performance regression shown in the second block in Table~\ref{tab:metrics_4s}. By contrast, the polynomial representation does not require extra regularization to produce physically realistic trajectories.
\section{Conclusion}
We proposed a simple and general trajectory representation that expresses continuous time-varying probabilistic predictions with polynomial parameterization. Detailed studies show that the polynomial representation is broadly effective, and can outperform the waypoint representation for low-count actors and large temporal supervision intervals. Moreover, we discussed the strength of the parametric representation in providing continuous and precise predictions, which is highly desired in applied robotics systems. Lastly, studying the physical feasibility of the predicted trajectories shows that the polynomial representation exhibits physical realism intrinsically without additional constraints or explicit regularization.



\bibliographystyle{IEEEtran}
\bibliography{references}

\begin{thebibliography}{10}
\providecommand{\url}[1]{#1}
\csname url@samestyle\endcsname
\providecommand{\newblock}{\relax}
\providecommand{\bibinfo}[2]{#2}
\providecommand{\BIBentrySTDinterwordspacing}{\spaceskip=0pt\relax}
\providecommand{\BIBentryALTinterwordstretchfactor}{4}
\providecommand{\BIBentryALTinterwordspacing}{\spaceskip=\fontdimen2\font plus
\BIBentryALTinterwordstretchfactor\fontdimen3\font minus
  \fontdimen4\font\relax}
\providecommand{\BIBforeignlanguage}[2]{{%
\expandafter\ifx\csname l@#1\endcsname\relax
\typeout{** WARNING: IEEEtran.bst: No hyphenation pattern has been}%
\typeout{** loaded for the language `#1'. Using the pattern for}%
\typeout{** the default language instead.}%
\else
\language=\csname l@#1\endcsname
\fi
#2}}
\providecommand{\BIBdecl}{\relax}
\BIBdecl

\bibitem{djuric2020}
N.~Djuric, V.~Radosavljevic, H.~Cui, T.~Nguyen, F.-C. Chou, T.-H. Lin, and
  J.~Schneider, ``Uncertainty-aware short-term motion prediction of traffic
  actors for autonomous driving,'' in \emph{IEEE Winter Conference on
  Applications of Computer Vision (WACV)}, 2020.

\bibitem{chen2017multi}
X.~Chen, H.~Ma, J.~Wan, B.~Li, and T.~Xia, ``Multi-view 3d object detection
  network for autonomous driving,'' in \emph{Proceedings of the IEEE Conference
  on Computer Vision and Pattern Recognition}, 2017, pp. 1907--1915.

\bibitem{casas2018intentnet}
S.~Casas, W.~Luo, and R.~Urtasun, ``Intentnet: Learning to predict intention
  from raw sensor data,'' in \emph{Conference on Robot Learning}, 2018, pp.
  947--956.

\bibitem{zhou2019end}
Y.~Zhou, P.~Sun, Y.~Zhang, D.~Anguelov, J.~Gao, T.~Ouyang, J.~Guo
  \emph{et~al.}, ``End-to-end multi-view fusion for 3d object detection in
  lidar point clouds,'' \emph{arXiv preprint arXiv:1910.06528}, 2019.

\bibitem{meyer2020laserflow}
G.~P. Meyer, J.~Charland, S.~Pandey, A.~Laddha, C.~Vallespi-Gonzalez, and C.~K.
  Wellington, ``Laserflow: Efficient and probabilistic object detection and
  motion forecasting,'' \emph{arXiv preprint arXiv:2003.05982}, 2020.

\bibitem{gao2020vectornet}
J.~Gao, C.~Sun, H.~Zhao, Y.~Shen, D.~Anguelov, C.~Li, and C.~Schmid,
  ``Vectornet: Encoding hd maps and agent dynamics from vectorized
  representation,'' in \emph{Proceedings of the IEEE/CVF Conference on Computer
  Vision and Pattern Recognition}, 2020, pp. 11\,525--11\,533.

\bibitem{luo2018fast}
W.~Luo, B.~Yang, and R.~Urtasun, ``Fast and furious: Real time end-to-end 3d
  detection, tracking and motion forecasting with a single convolutional net,''
  in \emph{Proc. of the IEEE CVPR}, 2018, pp. 3569--3577.

\bibitem{zhang2020stinet}
Z.~Zhang, J.~Gao, J.~Mao, Y.~Liu, D.~Anguelov, and C.~Li, ``Stinet:
  Spatio-temporal-interactive network for pedestrian detection and trajectory
  prediction,'' 2020.

\bibitem{djuric2020multixnet}
N.~Djuric, H.~Cui, Z.~Su, S.~Wu, H.~Wang, F.-C. Chou, L.~S. Martin, S.~Feng,
  R.~Hu, Y.~Xu, A.~Dayan, S.~Zhang, B.~C. Becker, G.~P. Meyer,
  C.~Vallespi-Gonzalez, and C.~K. Wellington, ``Multixnet: Multiclass
  multistage multimodal motion prediction,'' 2020.

\bibitem{ivanovic2019trajectron}
B.~Ivanovic and M.~Pavone, ``The trajectron: Probabilistic multi-agent
  trajectory modeling with dynamic spatiotemporal graphs,'' in
  \emph{Proceedings of the IEEE International Conference on Computer Vision},
  2019, pp. 2375--2384.

\bibitem{cui2019multimodal}
H.~Cui, V.~Radosavljevic, F.-C. Chou, T.-H. Lin, T.~Nguyen, T.-K. Huang,
  J.~Schneider, and N.~Djuric, ``Multimodal trajectory predictions for
  autonomous driving using deep convolutional networks,'' in \emph{2019
  International Conference on Robotics and Automation (ICRA)}.\hskip 1em plus
  0.5em minus 0.4em\relax IEEE, 2019, pp. 2090--2096.

\bibitem{zhao2020tnt}
H.~Zhao, J.~Gao, T.~Lan, C.~Sun, B.~Sapp, B.~Varadarajan, Y.~Shen, Y.~Shen,
  Y.~Chai, C.~Schmid, C.~Li, and D.~Anguelov, ``Tnt: Target-driven trajectory
  prediction,'' 2020.

\bibitem{alahi2016social}
A.~Alahi, K.~Goel, V.~Ramanathan, A.~Robicquet, L.~Fei-Fei, and S.~Savarese,
  ``Social lstm: Human trajectory prediction in crowded spaces,'' in
  \emph{Proceedings of the IEEE conference on computer vision and pattern
  recognition}, 2016, pp. 961--971.

\bibitem{cui2020deep}
H.~Cui, T.~Nguyen, F.-C. Chou, T.-H. Lin, J.~Schneider, D.~Bradley, and
  N.~Djuric, ``Deep kinematic models for kinematically feasible vehicle
  trajectory predictions,'' in \emph{2020 IEEE International Conference on
  Robotics and Automation (ICRA)}.\hskip 1em plus 0.5em minus 0.4em\relax IEEE,
  2020, pp. 10\,563--10\,569.

\bibitem{xu2014motion}
W.~Xu, J.~Pan, J.~Wei, and J.~M. Dolan, ``Motion planning under uncertainty for
  on-road autonomous driving,'' in \emph{2014 IEEE International Conference on
  Robotics and Automation (ICRA)}.\hskip 1em plus 0.5em minus 0.4em\relax IEEE,
  2014, pp. 2507--2512.

\bibitem{raksincharoensak2016motion}
P.~Raksincharoensak, T.~Hasegawa, and M.~Nagai, ``Motion planning and control
  of autonomous driving intelligence system based on risk potential
  optimization framework,'' \emph{International Journal of Automotive
  Engineering}, vol.~7, no. AVEC14, pp. 53--60, 2016.

\bibitem{kim2017probabilistic}
B.~Kim, C.~M. Kang, S.~H. Lee, H.~Chae, J.~Kim, C.~C. Chung, and J.~W. Choi,
  ``Probabilistic vehicle trajectory prediction over occupancy grid map via
  recurrent neural network,'' \emph{arXiv preprint arXiv:1704.07049}, 2017.

\bibitem{xue2018ss}
H.~Xue, D.~Q. Huynh, and M.~Reynolds, ``Ss-lstm: A hierarchical lstm model for
  pedestrian trajectory prediction,'' in \emph{2018 IEEE Winter Conference on
  Applications of Computer Vision (WACV)}.\hskip 1em plus 0.5em minus
  0.4em\relax IEEE, 2018, pp. 1186--1194.

\bibitem{Oh_2020_CVPR}
G.~Oh and J.-S. Valois, ``Hcnaf: Hyper-conditioned neural autoregressive flow
  and its application for probabilistic occupancy map forecasting,'' in
  \emph{Proceedings of the IEEE/CVF Conference on Computer Vision and Pattern
  Recognition (CVPR)}, June 2020.

\bibitem{Cui2020LearningDR}
Q.~Cui, H.~Sun, and F.~Yang, ``Learning dynamic relationships for 3d human
  motion prediction,'' in \emph{Proceedings of the IEEE/CVF Conference on
  Computer Vision and Pattern Recognition}, 2020, pp. 6519--6527.

\bibitem{li2020dynamic}
M.~Li, S.~Chen, Y.~Zhao, Y.~Zhang, Y.~Wang, and Q.~Tian, ``Dynamic multiscale
  graph neural networks for 3d skeleton based human motion prediction,'' in
  \emph{Proceedings of the IEEE/CVF Conference on Computer Vision and Pattern
  Recognition}, 2020, pp. 214--223.

\bibitem{tang2019adaptive}
\BIBentryALTinterwordspacing
C.~Tang, J.~Chen, and M.~Tomizuka, ``Adaptive probabilistic vehicle trajectory
  prediction through physically feasible bayesian recurrent neural network,''
  \emph{2019 International Conference on Robotics and Automation (ICRA)}, May
  2019. [Online]. Available: \url{http://dx.doi.org/10.1109/ICRA.2019.8794130}
\BIBentrySTDinterwordspacing

\bibitem{whiting1968empirical}
E.~E. Whiting, ``An empirical approximation to the voigt profile,''
  \emph{Journal of Quantitative Spectroscopy and Radiative Transfer}, vol.~8,
  no.~6, pp. 1379--1384, 1968.

\bibitem{su2018prl}
Z.~Su, A.~Zarassi, J.-F. Hsu, P.~San-Jose, E.~Prada, R.~Aguado, E.~J.~H. Lee,
  S.~Gazibegovic, R.~L.~M. Op~het Veld, D.~Car, S.~R. Plissard, M.~Hocevar,
  M.~Pendharkar, J.~S. Lee, J.~A. Logan, C.~J. Palmstr\o{}m, E.~P. A.~M.
  Bakkers, and S.~M. Frolov, ``Mirage andreev spectra generated by mesoscopic
  leads in nanowire quantum dots,'' \emph{Phys. Rev. Lett.}, vol. 121, p.
  127705, Sep 2018.

\bibitem{su2020prb}
Z.~Su, R.~\ifmmode~\check{Z}\else \v{Z}\fi{}itko, P.~Zhang, H.~Wu, D.~Car,
  S.~R. Plissard, S.~Gazibegovic, G.~Badawy, M.~Hocevar, J.~Chen, E.~P. A.~M.
  Bakkers, and S.~M. Frolov, ``Erasing odd-parity states in semiconductor
  quantum dots coupled to superconductors,'' \emph{Phys. Rev. B}, vol. 101, p.
  235315, Jun 2020.

\bibitem{baird1995residual}
L.~Baird, ``Residual algorithms: Reinforcement learning with function
  approximation,'' in \emph{Machine Learning Proceedings 1995}.\hskip 1em plus
  0.5em minus 0.4em\relax Elsevier, 1995, pp. 30--37.

\bibitem{gordon1995stable}
G.~J. Gordon, ``Stable function approximation in dynamic programming,'' in
  \emph{Machine Learning Proceedings 1995}.\hskip 1em plus 0.5em minus
  0.4em\relax Elsevier, 1995, pp. 261--268.

\bibitem{sutton2000policy}
R.~S. Sutton, D.~A. McAllester, S.~P. Singh, and Y.~Mansour, ``Policy gradient
  methods for reinforcement learning with function approximation,'' in
  \emph{Advances in neural information processing systems}, 2000, pp.
  1057--1063.

\bibitem{weierstrass1988}
H.~Jeffreys and B.~S. Jeffreys, \emph{Methods of Mathematical Physics}.\hskip
  1em plus 0.5em minus 0.4em\relax Cambridge University Press, 1988, pp.
  446--448.

\bibitem{lee2017desire}
N.~Lee, W.~Choi, P.~Vernaza, C.~B. Choy, P.~H. Torr, and M.~Chandraker,
  ``Desire: Distant future prediction in dynamic scenes with interacting
  agents,'' in \emph{Proceedings of the IEEE Conference on Computer Vision and
  Pattern Recognition}, 2017, pp. 336--345.

\bibitem{salzmann2020trajectron++}
T.~Salzmann, B.~Ivanovic, P.~Chakravarty, and M.~Pavone, ``Trajectron++:
  Multi-agent generative trajectory forecasting with heterogeneous data for
  control,'' \emph{arXiv preprint arXiv:2001.03093}, 2020.

\bibitem{gupta2018social}
A.~Gupta, J.~Johnson, L.~Fei-Fei, S.~Savarese, and A.~Alahi, ``Social gan:
  Socially acceptable trajectories with generative adversarial networks,'' in
  \emph{Proceedings of the IEEE Conference on Computer Vision and Pattern
  Recognition}, 2018, pp. 2255--2264.

\bibitem{zhao2019multi}
T.~Zhao, Y.~Xu, M.~Monfort, W.~Choi, C.~Baker, Y.~Zhao, Y.~Wang, and Y.~N. Wu,
  ``Multi-agent tensor fusion for contextual trajectory prediction,'' in
  \emph{Proceedings of the IEEE Conference on Computer Vision and Pattern
  Recognition}, 2019, pp. 12\,126--12\,134.

\bibitem{sadeghian2019sophie}
A.~Sadeghian, V.~Kosaraju, A.~Sadeghian, N.~Hirose, H.~Rezatofighi, and
  S.~Savarese, ``Sophie: An attentive gan for predicting paths compliant to
  social and physical constraints,'' in \emph{Proceedings of the IEEE
  Conference on Computer Vision and Pattern Recognition}, 2019, pp. 1349--1358.

\bibitem{kosaraju2019social}
V.~Kosaraju, A.~Sadeghian, R.~Mart{\'\i}n-Mart{\'\i}n, I.~Reid, H.~Rezatofighi,
  and S.~Savarese, ``Social-bigat: Multimodal trajectory forecasting using
  bicycle-gan and graph attention networks,'' in \emph{Advances in Neural
  Information Processing Systems}, 2019, pp. 137--146.

\bibitem{kong2015kinematic}
J.~Kong, M.~Pfeiffer, G.~Schildbach, and F.~Borrelli, ``Kinematic and dynamic
  vehicle models for autonomous driving control design,'' in \emph{2015 IEEE
  Intelligent Vehicles Symposium (IV)}.\hskip 1em plus 0.5em minus 0.4em\relax
  IEEE, 2015, pp. 1094--1099.

\bibitem{caesar2020nuscenes}
H.~Caesar, V.~Bankiti, A.~H. Lang, S.~Vora, V.~E. Liong, Q.~Xu, A.~Krishnan,
  Y.~Pan, G.~Baldan, and O.~Beijbom, ``nuscenes: A multimodal dataset for
  autonomous driving,'' in \emph{Proceedings of the IEEE/CVF Conference on
  Computer Vision and Pattern Recognition}, 2020, pp. 11\,621--11\,631.

\end{thebibliography}

\end{document}